\def\year{2022}\relax
\documentclass[letterpaper]{article} 
\usepackage{aaai22}  
\usepackage{times}  
\usepackage{helvet}  
\usepackage{courier}  
\usepackage[hyphens]{url}  
\usepackage{graphicx} 
\usepackage{natbib}  
\usepackage{caption} 
\DeclareCaptionStyle{ruled}{labelfont=normalfont,labelsep=colon,strut=off} 
\usepackage{algorithmic}

\usepackage{newfloat}
\usepackage{listings}
\usepackage[utf8]{inputenc} 
\usepackage[T1]{fontenc}    
\usepackage{times}
\usepackage{soul}
\usepackage{url}

\usepackage[utf8]{inputenc}
\usepackage{graphicx}
\usepackage{amsthm}
\usepackage{booktabs}
\usepackage{comment}
\usepackage[ruled,vlined, linesnumbered]{algorithm2e}
\usepackage{mathtools}
\usepackage{xcolor}
\usepackage{multirow}
\usepackage{booktabs}
\usepackage{colortbl}
\usepackage{subfig}
\usepackage[hidelinks]{hyperref}
\usepackage{authblk}

\usepackage[normalem]{ulem}
\useunder{\uline}{\ul}{}

\DeclareMathOperator*{\argmin}{\textbf{arg\,min}}

\newtheorem{theorem}{Theorem}

\title{Understanding and Measuring Robustness of Multimodal Learning}

\author[$^*$]{Nishant Vishwamitra}
\author[$^*$]{Hongxin Hu}
\author[$^*$]{Ziming Zhao}
\author[$^\dagger$]{Long Cheng}
\author[$^\dagger$]{Feng Luo}
\affil[$^*$]{Computer Science and Engineering, University at Buffalo}
\affil[$^\dagger$]{School of Computing, Clemson University}
\affil[ ]{$^*$\{nvishwam, hongxinh, zimingzh\}@buffalo.edu, $^\dagger$\{lcheng2, luofeng\}@clemson.edu}

\begin{document}
\nocopyright
\maketitle

\begin{abstract}
The modern digital world is increasingly becoming multimodal. Although multimodal learning has recently revolutionized the state-of-the-art performance in multimodal tasks, relatively little is known about the robustness of multimodal learning in an adversarial setting. In this paper, we introduce a comprehensive measurement of the adversarial robustness of multimodal learning by focusing on the \emph{fusion} of input modalities in multimodal models, via a framework called MUROAN (MUltimodal RObustness ANalyzer). We first present a unified view of multimodal models in MUROAN and identify the fusion mechanism of multimodal models as a key vulnerability. We then introduce a new type of multimodal adversarial attacks  called decoupling attack in MUROAN that aims to compromise multimodal models by decoupling their fused modalities.  We leverage the decoupling attack of MUROAN to measure several state-of-the-art multimodal models and find that the multimodal fusion mechanism in all these models is vulnerable to decoupling attacks. We especially demonstrate that, in the worst case, the decoupling attack of MUROAN achieves an attack success rate of 100\% by decoupling just 1.16\% of the input space. Finally, we show that traditional adversarial training is insufficient to improve the robustness of multimodal models  with respect to decoupling attacks. We hope our findings encourage researchers to pursue improving the robustness of multimodal learning.
\end{abstract}

\section{Introduction}
\label{sec:introduction}

Multimodal learning has been gradually gaining focus of the research community over the past few years. The approaches for multimodal learning have come a long way from simple models re-purposed for multimodal tasks, to deep learning-based models that are specifically designed for multimodal tasks (referred to as Deep Multimodal Models or DMMs throughout this paper). For example, recent advances in this field have led to several state-of-the-art DMMs, such as ViLBERT~\cite{lu2019vilbert}, VisualBERT~\cite{li2019visualbert}, MMBT~\cite{kiela2019supervised}, and Pythia~\cite{jiang2018pythia}, while also engendering the collection of several multimodal datasets, such as Hateful Memes~\cite{kiela2020hateful}, Visual Question Answering (VQA)~\cite{balanced_vqa_v2}, and Visual Commonsense Reasoning (VCR)~\cite{zellers2019recognition}. Due to the success of these DMMs on standard benchmarks, there have been many encouraging attempts to adopt them to real-world and safety-critical scenarios, such as assistance to blind people~\cite{gurari2018vizwiz},  hate-speech moderation on social media~\cite{kiela2020hateful}, as well as emerging domains, such as Google MUM search~\cite{googlemum}. However, in spite of the recent advances, the robustness of DMMs remains poorly understood.

A significant difference between DMMs and their unimodal counterparts is the \textit{fusion} mechanism in DMMs. This fusion mechanism fuses multiple input modalities to learn their joint representation, which is then processed by several fully connected layers to predict classification scores depending on the nature of the corresponding downstream tasks. Different DMMs~\cite{lu2019vilbert, kiela2019supervised, li2019visualbert, jiang2018pythia} employ different strategies to learn strong fusion embeddings of their input modalities. This fusion mechanism presents new challenges towards studying the adversarial robustness of these models. 

\begin{figure}[t]
  \centering
    \resizebox{0.5\textwidth}{!}{\includegraphics[width=1.0\linewidth, scale=1.0]{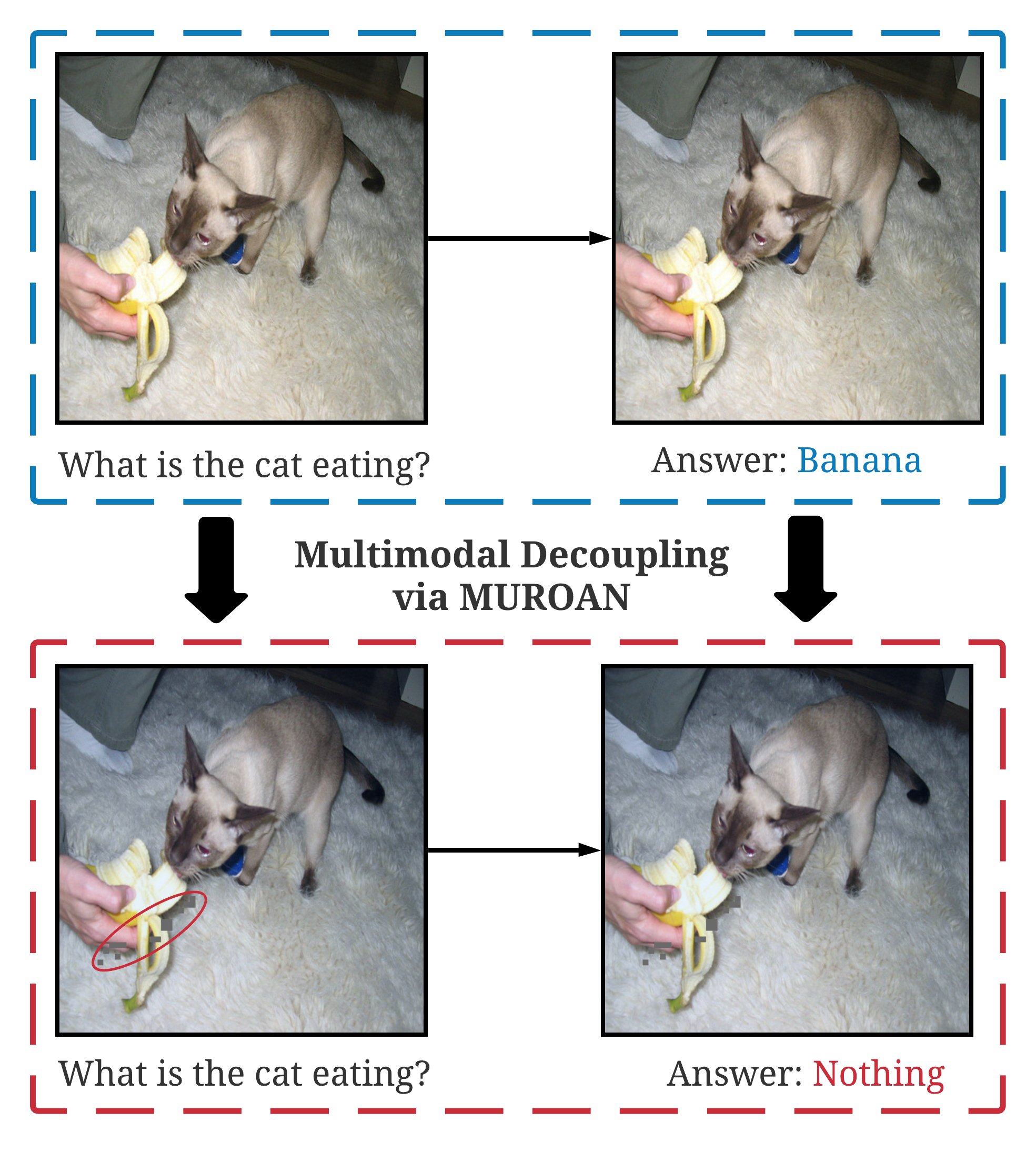}}
\caption{By decoupling the input modalities through removal of a few datapoints in the image via MUROAN framework, the multimodal model predicts a wrong answer class: \textit{Nothing}, indicating that decoupling attack can easily compromise multimodal models.}
\label{fig:intro}
\end{figure}

Recently, several unimodal adversarial attacks for deep unimodal models have been formulated to study their robustness. For example, unimodal adversarial images~\cite{szegedy2013intriguing, madry2017towards, papernot2016limitations, wicker2018feature, carlini2017towards, wang2020t3} and unimodal adversarial text~\cite{alzantot2018generating, li2018textbugger, jin2019bert, alzantot2018generating, ren2019generating} have been widely studied, which have exposed numerous vulnerabilities in the deep unimodal models. However, these attacks cannot be directly employed to study the robustness of their deep multimodal counterparts. First, since these attacks can only be applied to single modalities, they do not affect the fusion mechanism that is fundamental to DMMs. 
Second, since DMMs combine several different types of modalities (e.g. image, text, speech, etc.), a single unimodal attack cannot be used for all those modalities. We note that formulating comprehensive methods to study the robustness of DMMs is of utmost importance to adopting them in real-world systems, such as VQA. 

To address these challenges, in this work, we first highlight how multimodal adversarial attacks based on decoupling the input modalities in DMMs can easily compromise these models. Then, we introduce a framework called MUROAN to study the robustness of DMMs based on decoupling of modalities, thereby revealing vulnerabilities in the fusion mechanism of existing DMMs. MUROAN uses a unified view of DMMs to expose its key vulnerability. Then, we introduce a new type of adversarial attacks called decoupling attack in MUROAN, wherein the objective of its attack algorithm is to decouple the input modalities of multimodal models to induce a misclassification. As depicted in Figure~\ref{fig:intro},  a decoupling of the image and text modalities through occlusion of a few datapoints in the image induces a misclassification. In addition, we leverage the MUROAN framework to measure  several state-of-the-art DMMs. We find that the seemingly straightforward decoupling attack of MUROAN is in fact highly effective in compromising DMMs.

Our contributions in this work are as follows.

\begin{itemize}
  \item We present a unified view of DMMs to explore their vulnerabilities, and identify the fusion mechanism of these models as a critical component for their robustness analysis. 
  
  \item We propose a novel framework called MUROAN that consists of the unified view to exploit the fusion mechanism and a decoupling attack algorithm for comprehensively studying the adversarial robustness of DMMs. MUROAN directly focuses on the fusion mechanism of DMMs by decoupling the input modalities that are fused together. 
  
  \item We use MUROAN for a comprehensive robustness analysis of state-of-the-art DMMs under several dataset and model settings. Our experiments show that, in the worst case, the decoupling attack in MUROAN can achieve an attack success rate of 100\% after decoupling of 1.16\% of input modalities of DMMs, while the unimodal adversarial attacks overestimate the robustness of DMMs. 
  
\end{itemize}

We are open-sourcing our code to encourage research in training DMMs robust to decoupling attacks: \url{http://github.com/SecurityAndPrivacyResearch/mda}.

\section{Background}
\label{sec:background}
In the following, we give an overview of the field of  multimodal learning as well as the state-of-the-art unimodal adversarial attacks used for the robustness analysis of unimodal models.

\subsection{Multimodal Learning} 
The renewed interest in multimodal learning can be attributed to more powerful models~\cite{devlin2018bert, vaswani2017attention} that can learn strong fusion of input modalities and the availability of several multimodal datasets~\cite{balanced_vqa_v2, zellers2019recognition, kiela2020hateful}. These models and datasets have resulted in DMMs achieving impressive results on standard benchmarks. Much of the DMMs that have achieved impressive performances can be categorized under the following categories.

\textbf{Traditional Fusion-based Models.} Several DMMs have attempted to address how to effectively combine multimodal information~\cite{baltruvsaitis2018multimodal,bruni2014multimodal,lazaridou2015combining}. Feature concatenation is one of the most preferred fusion techniques in these models, while some of the models use other feature fusion techniques such as element-wise product. Since these models showed impressive performances on several multimodal benchmarks, they are considered strong baselines for many multimodal tasks.

\textbf{Transformer-based Fusion Models.} Recently, the BERT model~\cite{devlin2018bert}, a type of transformer~\cite{vaswani2017attention}, has been shown to achieve state-of-the-art performance~\cite{kiela2019supervised,li2019visualbert,su2019vl} on multimodal benchmarks, by learning the interaction between the input modalities via self-attention over many different layers. For example the MMBT~\cite{kiela2019supervised} model fuses image embeddings in the form of pooled filter maps from a ResNet model and word tokens as two segments of BERT~\cite{devlin2018bert}. Similarly, the VL-BERT~\cite{su2019vl} model fuses regions of interest (ROIs) of an image with word tokens as two segments of BERT. As shown by these works, the transformer based DMMs outperform their unimodal counterparts in multimodal tasks by quite a large margin.

\subsection{Unimodal Adversarial Attacks} 
The discovery of unimodal adversarial attacks has engendered active research in the safety and robustness of unimodal deep learning models. In this section, we discuss important unimodal adversarial attacks on images and text.

\textbf{Unimodal Adversarial Image.} A large body of adversarial attacks have been introduced in recent times that mainly focus towards robustness analysis of computer vision models. For example, several works, such as fast-gradient attacks~\cite{goodfellow2014explaining,liu2016delving}, optimization-based methods~\cite{szegedy2013intriguing,carlini2017towards}, and other such methods~\cite{papernot2016limitations,nguyen2015deep}, have been proposed successfully. Recently, ensemble based attacks~\cite{croce2020reliable} have been show to more deeply reveal vulnerabilities in unimodal models.  Furthermore, alarmingly critical real-world attacks such as adversarial patches~\cite{brown2017adversarial} have been introduced recently, which cast serious questions on the safety of these vision models.

\textbf{Unimodal Adversarial Text.} Recently, some works have focused on unimodal adversarial text to study robustness of Natural Language Processing (NLP) models. While earlier works~\cite{li2018textbugger, gao2018black, eger2019text} effectively employed character level perturbations to perform adversarial attacks, more recent works have found word replacement strategies~\cite{jin2019bert, alzantot2018generating, ren2019generating} to be largely effective in compromising these models. Recent works ~\cite{iyyer2018adversarial, zhao2017generating, ribeiro2018semantically} have also demonstrated how sentences can be merely reconfigured to pose serious adversarial threats.


Recently, some studies have emerged that discuss adversarial attacks on DMMs~\cite{tian2021can, li2020vulnerability}. However, these studies do not focus on exploring the vulnerabilities of the fusion mechanism to adversarial attacks. In this work, we specifically focus on comprehensively studying the adversarial robustness of DMMs via a type of multimodal adversarial attack called decoupling attack, that focuses on decoupling the input modalities of a DMM to compromise the fusion mechanism of these DMMs.

\section{Decoupling  Input Modalities}
\label{sec:prelim_analysis_unimodal_attacks}
Our primary objective in this section is to demonstrate how easily decoupling of input modalities can compromise DMMs. 
To this end, we performed a preliminary experiment for comparing the effect of decoupling attacks on DMMs against traditional unimodal adversarial attacks.

We randomly selected 100 samples from the VQA dataset~\cite{antol2015vqa} and the pretrained Pythia DMM~\cite{jiang2018pythia} to conduct our preliminary study. Several previous unimodal attacks~\cite{goodfellow2014explaining,madry2017towards, kurakin2016adversarial, moosavi2016deepfool, papernot2016limitations, xie2019improving, dong2018boosting} have revealed the nature of different vulnerabilities in traditional unimodal model. In this experiment, we use the state-of-the-art attack called PGD attack~\cite{madry2017towards} as the traditional, unimodal adversarial attacks. Next, we manually studied the 100 samples and occluded datapoints that we considered as participating in the fusion mechanism. Our objective from this step was to manually decouple the DMMs to study whether decoupling could be considered as an effective means to create adversarial attacks that can be comparable against strong and popular unimodal attacks in terms of their effectiveness in fooling the DMM.

We found that just manual decoupling was able to effectively fool 50\% of the samples considered in the experiment. But more importantly, we found that on average, manual decoupling only affected 7.25\% of the datapoints in the image of each sample. On the other hand, we found that although the PGD attack was quite effective in compromising the DMM with close to 100\% success rate, 96.47\% of the datapoints on average were affected by PGD. What this experiment shows is that unimodal adversarial attacks are not able to identify the optimum datapoints to perturb. Thus, unimodal attacks are not sufficiently suitable for studying robustness of DMMs. Furthermore, since unimodal attacks do not seem to take the fusion mechanism into consideration, they do not reveal the vulnerabilities specific to DMMs. In the sections that follow, we show how MUROAN decoupling attack algorithm can optimally find the exact datapoints involved in fusion, so that the adversarial robustness through decoupling can be studied.




\section{Robustness Analysis}

\begin{figure*}[t]
  \centering
    \resizebox{0.9\textwidth}{!}{\includegraphics[width=1.0\linewidth, scale=1.2]{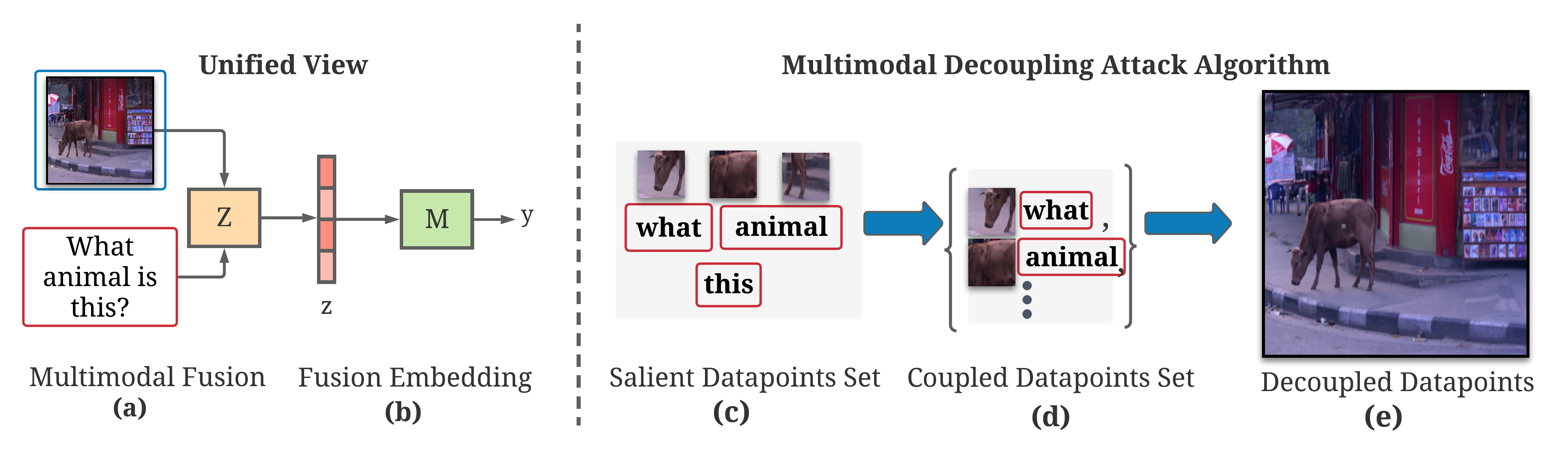}}
\caption{Overview of our approach.}
\label{fig:overview}
\end{figure*}

In this section, we discuss our approach for the robustness analysis of DMMs via MUROAN framework. In this regard, we first discuss a unified view of DMMs to explore the vulnerabilities of the fusion mechanism of DMMs, and then introduce our algorithm to decouple the fused modalities of DMMs. The overview of our approach is depicted in Figure~\ref{fig:overview}.

\subsection{Unified View of Deep Multimodal Models}
We consider a DMM $D : X \rightarrow Y$ to be a function that maps a domain $X$ to a co-domain $Y$. An input is a set of vectors of different modalities $x = \{x^1_0 \dots x^1_n, \, x^2_0 \dots x^2_m, \, \dots\}$ (Figure~\ref{fig:overview}, Step (a)). We consider $Y$ to be the set of possible classes for a multimodal input $x \in X$. The output of the DMM for a multimodal input $x$ is considered to be $D(x) = y$, for some $y \in Y$. We denote the confidence of the DMM for a multimodal classification probability on input $x$ and class $y$ as $D_y(x)$. Lastly, we denote the cardinality of a set as $|\cdot|$, which represents the number of elements in the set.

Although DMMs have several different architectural configurations, we need a unified view (or representation) of them for a uniform vulnerability analysis of all these different multimodal architectures. To achieve this, we unify these different architectural approaches into a single view, in which we consider a DMM as a generator of the fusion embedding of multiple input modalities (Figure~\ref{fig:overview}, Step (b)), followed by several fully connected layers that are specific for downstream tasks. In other words, we break down a DMM into two functions: the first generates a latent representation (i.e., the fusion embedding) of the multimodal inputs and the second performs classification based on the fusion embedding. We consider the fusion embedding of a multimodal input $x$ as $Z(x) = z$, where $z$ is the $d$-dimensional fusion embedding vector. Next, we consider $y = M(z)$ to represent classification based on the fusion embedding from fully connected layers that are specific to downstream tasks. Therefore, the original DMM is broken down into two functions, represented as $M(Z(x))$. We further discuss this process for two typical DMM architectures: traditional architectures and transformer-based architectures.

\textbf{Traditional Multimodal Architectures.} Traditional DMM architectures are composed of separate neural networks that are specific to each input modality, whose outputs are combined using fusion techniques such as element-wise multiplication, addition or concatenation. For example, the Pythia~\cite{jiang2018pythia} architecture is composed of a convolutional neural network that learns the embedding of the image modality, and a recurrent network that learns the embedding of the text modality, which are then combined using element-wise multiplication. This combination represents the fusion embedding.

\textbf{Transformer-based Multimodal Architectures.} These architectures use the transformer~\cite{vaswani2017attention} for learning a strong fusion embedding of the input modalities. The input modalities are first converted into embeddings, which are then combined using the transformer, which performs several self-attentions across many layers. The first token embedding then constitutes the fusion embedding, which is subsequently processed by fully connected layers for classification. 

\subsection{MUROAN Framework}
\label{sec:approach}
We note that the traditional methods of adversarial attacks are not suitable for DMMs for two specific reasons. First, most key methods of crafting adversarial attacks use either the $l_\infty$ or $l_2$ norm. Optimization with respect to these kinds of manipulations induces a perturbation in all (or almost all) of the datapoints of an input modality by a small value $\pm\epsilon$. This is not suitable in case of multimodal inputs because different modalities have different compositions, and not all modalities support this type of manipulation. For example, image-based inputs are continuous and thus suitable for such manipulations, but text-based inputs are discrete, thus not suitable for such manipulations. Furthermore, for DMMs, such adversarial manipulations are not suitable for robustness analysis processes since the core weaknesses of these models should be examined in the fusion mechanism of these models, which is not achieved by these manipulations. Since we are interested in studying the effect of decoupling fused modalities, we employ $l_0$-norm optimization attack algorithm, wherein an $l_0$-norm attack optimizes for the number of changes made to the inputs for a successful decoupling attack. 

Removal of salient datapoints from inputs has been shown to be an important factor for considering the robustness and safety of a decision model~\cite{mathias2013handling, wicker2019robustness, noh2018improving}. However, the key difference between the traditional unimdoal domains and the multimodal domain is that such datapoints are in fact parts of separate modalities that are coupled together by the multimodal fusion mechanism. Thus, it is imperative to study the cases, in which some parts of the input modalities are removed, so as to render this fusion as unsuccessful.

For a multimodal input $x$, we consider coupled datapoints as some $x^\prime \subset x$. Our objective is to find the minimum subset via the following optimization.

\begin{equation}
\label{eq:optimization}
\argmin_{x^\prime \subset x}(|x| - |x^\prime|) \, \ni \, D(x) \neq D(x^\prime)
\end{equation}

However, it is impractical to solve the optimization in Equation~\ref{eq:optimization}, due to a large number of such datapoints in the multimodal input space. Thus, to solve this optimization, we use the notion of the fusion embedding to compute a salient points set first, $S_n$ (Figure~\ref{fig:overview}, Step (c)). Such sets of critical or salient points have been previously utilized to inspect deep learning-based  models~\cite{qi2017pointnet}. We use the salient datapoints set to study the weaknesses of DMMs, by defining it as follows.

\begin{equation}
\label{eq:salient_set}
S_n^x = \{x_i \in x \, | \, Z(x/x_i) \neq Z(x)\}
\end{equation}

In Equation~\ref{eq:salient_set}, the salient datapoints set contains those datapoints that affect the fusion embedding upon removal (where $x/x_i$ denotes removal of a datapoint). For example in the transformer-based DMMs, a datapoint $x_i \in S_n^x$ if $\forall i \neq j,z_i \geq z_j$ due to the transformer pooling layer. Next, we find the set of coupled datapoints (Figure~\ref{fig:overview}, Step (d)) from the salient datapoints set, by computing permutations of all datapoints of a maximum size equal to the size of the salient datapoints set (denoted by $\prod$ in Equation~\ref{eq:coupled_set}).

\begin{equation}
\label{eq:coupled_set}
C_n^x = \prod_{i = 1}^{||S_n^x||} \{s_i\}
\end{equation}

\begin{algorithm}[t]
\SetAlgoNoLine
\KwIn{x, y, D, $\Theta$, f, \textbf{\text{maxitr}}}

\KwOut{$x^\prime$}

Initialization: $x^\prime \leftarrow x$

\While{ f(D, x, $x^\prime$, y) \textup{ or } \textup{\textbf{maxitr}}}{

  $S_n^x \leftarrow \textup{GetSalientSet}(D, x^\prime)$
  
  $C_n^x \leftarrow \textup{GetCoupledSet}(S_n^x)$
  
  \For{$x_i \in C_n^x$}{
    
    \If{$D(x^\prime) \neq y$}{
     
     break
     
    }
    
    $x^\prime \leftarrow x^\prime / x_i$
    
    \If{$D(x^\prime) \neq y \textup{ and } f(D, x, x^\prime, y) \neq \textup{True}$}{
    
      $x \leftarrow x$
    
    }
    
  }

}
\Return $x^\prime$
\caption{MUROAN Decoupling Attack Algorithm}
\label{alg:mm_decoupling_attack_algorithm}
\end{algorithm}

Now that we have computed the coupled datapoints set, we propose the MUROAN Decoupling Attack Algorithm (Algorithm~\ref{alg:mm_decoupling_attack_algorithm}) to iteratively refine the decoupling attack. In our algorithm, first the salient datapoints set is computed based on the process described in Equation~\ref{eq:salient_set}. Then, the GetCoupledSet procedure is called, which performs two functions. First, the coupled datapoints are computed as described in Equation~\ref{eq:coupled_set}. Then, they are ordered based on the size of the datapoints, so as to satisfy Equation~\ref{eq:optimization}. We encode the termination of our algorithm as a boolean function $f$, to support multiple adversarial requirements. For example, adversarial requirements for crafting untargeted attacks ($D(x^\prime) \neq y$) or targeted attacks ($D(x^\prime) = y^\prime$) can be supported (Figure~\ref{fig:overview}, Step~(e)).
Lastly, we propose the following theorem to use our decoupling attack algorithm as a robustness verification technique to find adversarial examples in DMMs if one exists.

\begin{theorem}
For a multimodal model $D$ that satisfies our unified view and a given multimodal input $x$, the MUROAN decoupling attack algorithm will find the optimum adversarial example that satisfies Equation~\ref{eq:optimization}.
\end{theorem}

\noindent\textit{Proof.} If an adversarial example exists for input $x$, it can be found by an exhaustive search of the input space. The GetCoupledSet function returns all possible permutations of the coupled datapoints and the $f$ function and \textbf{maxitr} can be set such that the algorithm does not terminate until a satisfactory adversarial permutation is found. Furthermore, since the permutations in the coupled datapoints set are ordered, thus, a permutation that is found by our algorithm to be adversarial is minimal.

\section{Experiments}
\label{sec:experiments}

In our evaluation, we use MUROAN to analyze the robustness of state-of-the-art DMMs trained on popular multimodal datasets to show how decoupling attack can easily compromise these models, thereby enabling us to understand their robustness. We also consider some unimodal adversarial attack baselines in our evaluation only to show how easily decoupling attack can compromise DMMs. Our objective is not to make a direct comparison of our approach against these existing attacks, but to highlight how decoupling of input modalities can be easily used to attack the fusion mechanism of DMMs. Our findings highlight the need for rigorous safety analysis of DMMs against decoupling attacks, and lay down important groundwork for their deployment in real-world applications. We first summarize the DMMs, datasets, and unimodal adversarial baselines that are used in our experiments.

\subsection{Deep Multimodal Models}
\label{sec:multimodal_models}

\begin{itemize}
  \item \textbf{Pythia}. The Pythia~\cite{jiang2018pythia} is a state-of-the-art model in the VQA challenge task. This models is composed of a convolutional network to compute an image embedding and a recurrent network to compute a sentence embedding, which are fused using element-wise multiplication. 
  
  \item \textbf{Late Fusion}. We consider the late-fusion architecture based DMM in~\cite{antol2015vqa} as a strong baseline model. In this model, image embeddings from a convolutional neural network and text embeddings from a recurrent network are fused using element-wise sum, and then the fusion embedding is processed through multiple classification layers to generate a probability score. 
  
  \item \textbf{MMBT}. The MMBT model~\cite{kiela2019supervised} is a state-of-the-art DMM that utilizes the BERT~\cite{devlin2018bert} to learn multimodal embeddings by the implicit alignment of image and text features with the self-attention mechanism of transformers~\cite{vaswani2017attention}, for a wide range of visual-linguistic tasks. The query vector of this model, which is treated as the fusion embedding, is processed through a classifier head for downstream tasks. 
\end{itemize}

\subsection{Multimodal Datasets}

\begin{itemize}
  \item \textbf{Hateful Memes}. The Hateful Memes~\cite{kiela2020hateful} dataset consists of image and text pairs pertaining to hateful memes, a recent phenomenon that poses a serious threat societal threat in today's day and age. The objective is classification into two categories: ``hateful'' or ``non-hateful''. 
  
  \item \textbf{Visual Question Answering (VQA)}. The VQA dataset~\cite{antol2015vqa} consists of images with multiple associate natural language questions. Each image and question pair expects a list of answers. The objective is to predict the best answer from the list of answers for each image-question pair.
  
\end{itemize}

\subsection{Unimodal Adversarial Baselines}

\begin{itemize}
  \item \textbf{CW Attack}. We use the Carlini and Wagner~\cite{carlini2017towards} attack algorithm as baseline for unimodal adversarial images for image-based modality.
  
  \item \textbf{Genetic Attack}. We use the Genetic Attack~\cite{alzantot2018generating} algorithm (referred to as ``Genetic'' in this paper) as baseline for unimodal adversarial text for text-based modality.
  
\end{itemize}

\subsection{Other Implementation Details}
\label{sec:impl_details}
We have implemented our attack using the PyTorch~\cite{NEURIPS2019_9015} library. For the VQA dataset we used 1000 samples and for Hateful Memes dataset, we used 250 samples to conduct our experiments. We used pretrained models~\cite{pretrained} published by the original authors for all the DMMs that we have evaluated in our experiments. In the MUROAN decoupling attack algorithm, we used a maximum iteration limit of 500 epochs, post which we report the attack as unsuccessful. We ran our experiments on a single NVIDIA V100 GPU enabled eight core machine.

\subsection{Robustness Analysis }
\label{sec:Robustness_Analysis}
\begin{figure}[t]
\centering
\includegraphics[width=0.9\columnwidth]{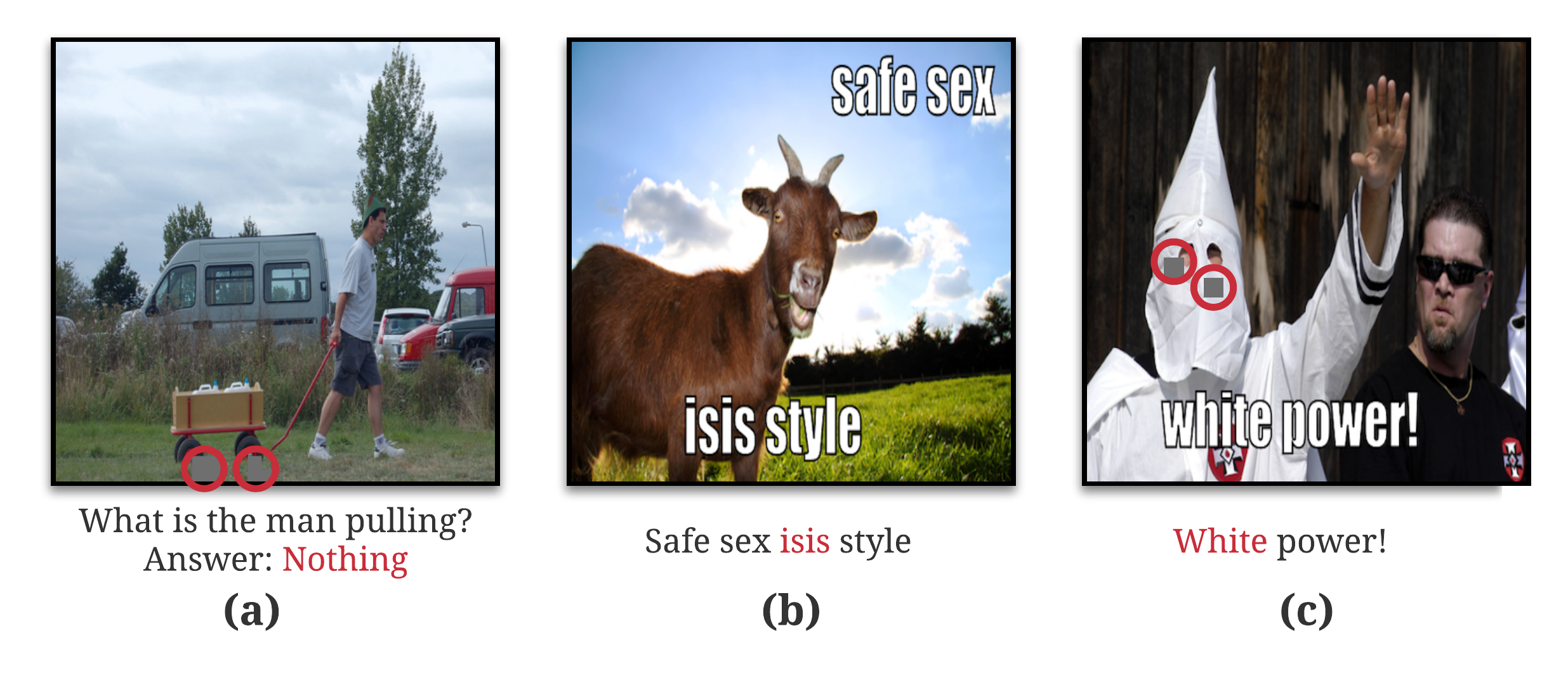} 
\caption{Three samples depict three types of minimum coupled datapoints in the VQA and Hateful Memes dataset. In sample (a), the minimum coupled datapoints are in the image only (indicated by red circles), and it is enough to only make changes to a those datapoints to decouple the sample. In sample (b), the minimum coupled datapoints are in the text only (indicated by red font), it is enough to make changes to the text only to decouple the sample. In sample (c), the coupled datapoints consist of both image and text, therefore both need to be changed to decouple this sample.}
\label{fig:adv_samples}
\end{figure}
In this section, we used our framework to analyze the robustness of state-of-the-art DMMs under various attack conditions to show that the robustness of these DMMs are largely overestimated.

\begin{figure}[tb!]
\vspace{-0.3cm}
    \centering
    \begin{subfloat}[][Pythia-VQA]{
        \includegraphics[width=0.7\linewidth]{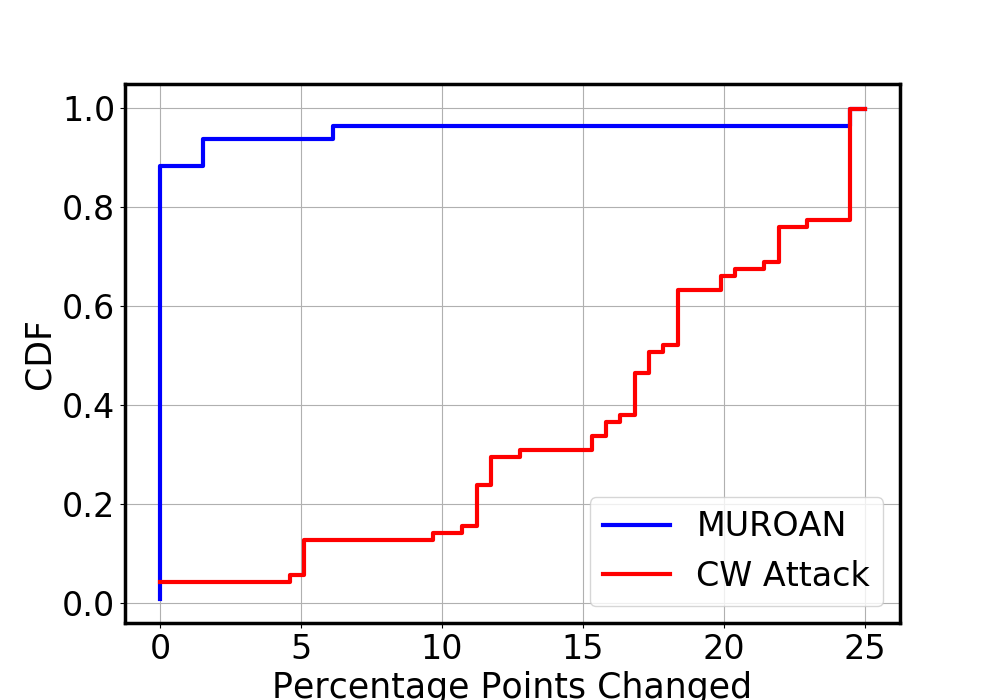}\label{fig:pythia_vqa}}
    \end{subfloat}
    ~%
    \begin{subfloat}[][Late Fusion-Hateful Memes]{
        \includegraphics[width=0.7\linewidth]{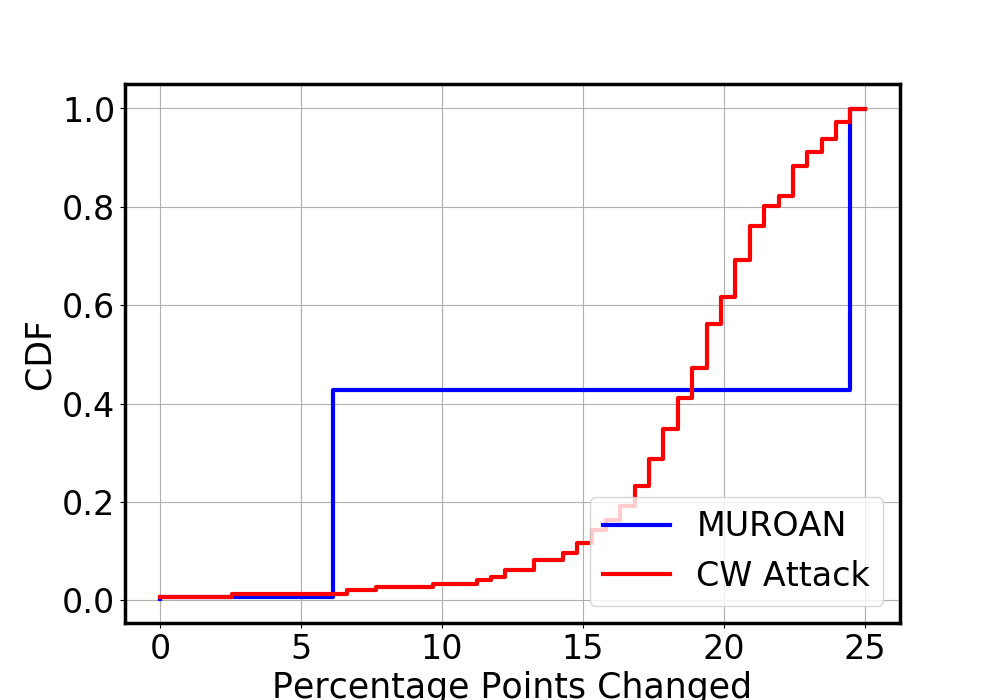}\label{fig:late_fusion_hateful_memes}}
    \end{subfloat}
    ~%
    \begin{subfloat}[][MMBT-Hateful Memes]{
        \includegraphics[width=0.7\linewidth]{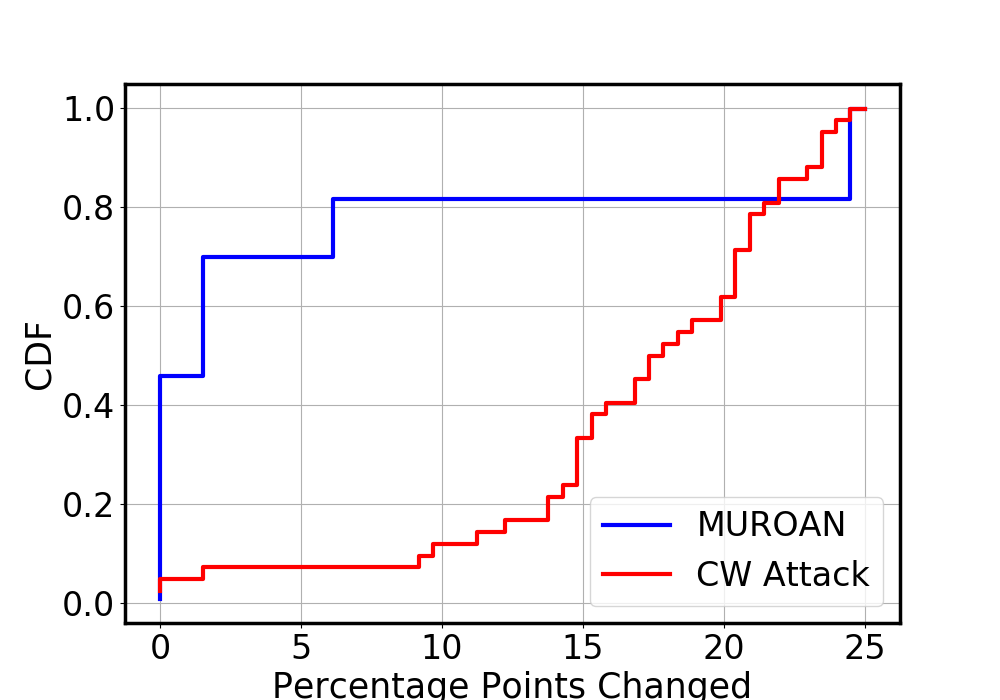}\label{fig:mmbt_hateful_memes}}
    \end{subfloat}
    \caption{CDF of percentage of datapoints changed. }
    \label{fig:cdf}
    \vspace{-0.5cm}
\end{figure}

We studied the percentage of points changed by MUROAN decoupling attack algorithm in comparison with the CW attack for a successful misclassification. We used the same cutoff of 500 epochs for both the algorithms in all the tests, post which we reported a failure. We have depicted the results of this experiment in Figure~\ref{fig:cdf}.

\begin{table*}[h]
\centering
\begin{tabular}{c|c|c}
\textbf{Model-Dataset}             & \textbf{Average Points Changed - MUROAN} & \textbf{Average Points Changed - CW} \\ \hline
Pythia-VQA                & \textbf{1.16\%}                                  & 93.99\%                                \\ \hline
Late Fusion-Hateful Memes & \textbf{16.93\%}                                 & 99.86\%                                \\ \hline
MMBT-Hateful Memes        & \textbf{5.73\%}                                  & 94.92\%                                \\ 
\end{tabular}
\caption{Comparison of Average Percentage Points Affected by MUROAN and CW attack.}
\label{tab:average_percentage_points_changed}
\end{table*}

Figure~\ref{fig:cdf} depicts the CDF of the average percentage of datapoints changed in both the attacks under consideration. We found that the unimodal adversarial images (i.e., the CW attack) vastly overestimated the robustness of all the three DMMs. For the Pythia-VQA, it was observed that the CW attack changed 93.99\% of the input datapoints, whereas MUROAN decoupling attack algorithm changed 1.16\% of input datapoints. This difference of a large margin showed that the baseline unimodal adversarial images vastly overestimated the robustness of models for the VQA task. This finding may have important implications on using VQA in real-world applications, such as visual question answering for the blind~\cite{gurari2018vizwiz}. Next, we discuss another important application domain, namely Hateful Memes. For the Late Fusion- Hateful Memes model, it was again observed that the CW attack changed 99.86\% of datapoints, whereas MUROAN decoupling attack algorithm changed an average of 16.93\% of input datapoints, a significant difference. For the MMBT-Hateful Memes model, it was observed that the CW attack changed 94.92\% of datapoints, whereas MUROAN decoupling attack algorithm changed an average of 5.73\% of input datapoints. In both cases, the unimodal adversarial images overestimated the robustness of the DMMs trained for the Hateful Memes task. 

\begin{table*}[h]
\centering
\begin{tabular}{c|c|c|c|c}
\textbf{Model-Dataset}    & \textbf{ASR-MUROAN} & \textbf{ASR-CW} & \textbf{ASR-Genetic} & \textbf{ASR-CW+Genetic} \\ \hline
Pythia-VQA                & \textbf{100\%}            & 79.77\%         & 49.30\%                     & 86.45\%                     \\ \hline
Late Fusion-Hateful Memes & \textbf{97.25\%}          & 59.11\%         & 0\%                     & 59.11\%                         \\ \hline
MMBT-Hateful Memes        & \textbf{83.33\%}          & 47.19\%         & 0\%                     & 47.19\%                         \\ 
\end{tabular}
\caption{Comparison of Attack Success Rate (ASR).}
\label{tab:asr}
\end{table*}

Next, we compared the Attack Success Rate (ASR) of MUROAN decoupling attack algorithm with respect to the unimodal adversarial images and text baselines, namely the CW~\cite{carlini2017towards} attack and the Genetic~\cite{alzantot2018generating} attack respectively. The results of this experiment have been depicted in Table~\ref{tab:asr}. We first discuss the impact of the unimodal adversarial images on the DMMs. In all the three DMMs, we found that the unimodal adversarial images could affect these DMMs. However, they vastly overestimated their robustness in all three cases, when we compared the ASRs of MUROAN decoupling attack algorithm. For the Pythia-VQA model, the CW attack achieved an ASR of 79.77\%, although the ASR achieved by MUROAN decoupling attack algorithm was 100\%. For the two DMMs for hateful memes (i.e., Late Fusion-Hateful Memes and MMBT-Hateful Memes), a similar observation was made, although the CW attack achieved significantly lower ASR for both DMMs. Next, we took a closer look at the impact of the unimodal adversarial text (i.e., Genetic attack) on the DMMs, in comparison with MUROAN. For the Pythia-VQA, it was observed that the Genetic attack has little effect when compared to MUROAN decoupling attack algorithm, and even to the CW attack, wherein both these attacks outperformed the unimodal adversarial text baseline by a large margin. In case of the hateful memes DMMs (i.e., Late Fusion-Hateful Memes and MMBT-Hateful Memes) this margin was found to be even larger. It was observed that the unimodal adversarial text had no significant effect on the DMMs for hateful memes. 

Thus, we observed that the safety and robustness of these DMMs need to be deeply examined, specifically from the perspective of decoupling attacks. In this regard, our experiments indicate that our attack exposes the vulnerabilities in the fusion mechanism of DMMs, and the robustness of this mechanism needs significant improvement, especially if DMMs are to be deployed in real-world systems.

\subsection{Qualitative Analysis of MUROAN}
\label{sec:qualitative_analysis_of_mda}




In this section, we provide a qualitative analysis of the decoupled samples that our the MUROAN decoupling attack algorithm generated. Upon observation of such samples in the two baseline datasets (i.e., VQA and Hateful Memes), we discuss certain aspects of the nature of decoupling pertaining to our observations. In Figure~\ref{fig:adv_samples}~\footnote{Note: samples (b) and (c) are from the Hateful Memes dataset~\cite{kiela2020hateful}, which some readers may find distressing.}, we depict three samples from our robustness analysis experiments. Figure~\ref{fig:adv_samples} (a) is from the VQA dataset, and Figures~\ref{fig:adv_samples} (b) and (c) are from the Hateful Memes dataset. These three samples represent the three levels of decoupling we observed in our experiments. In Figure~\ref{fig:adv_samples} (a), the minimum coupled datapoints were found in the image only, therefore it is sufficient to decouple just the single image modality. In the VQA dataset, since questions are asked about certain parts of an image, this observation is intuitive since it should be sufficient to only affect the relevant parts of the image. In Figure~\ref{fig:adv_samples} (b), the minimum coupled datapoints were only found in the text modality, since intuitively we cannot see why this sample could be a hateful meme from the image alone. In Figure~\ref{fig:adv_samples} (c), the minimum coupled datapoints consist of both the image and the text modalities. In this case, both the input modalities need to be affected for decoupling this fusion. Therefore, we note that vulnerabilities in the DMMs are of a  very different nature when compared to their unimodal counterparts. 

\subsection{Adversarial Training}
\label{sec:adversarial_training}
Our experiments in Section~\ref{sec:Robustness_Analysis} raise an important question: how can we defend against decoupling attacks? We performed a preliminary experiment to see if 
adversarial training~\cite{goodfellow2014explaining}, a popular technique to improve adversarial robustness, can be used to reduce the attack success rate.  We performed adversarial training using the MMBT model for the hateful memes classification. We generated 247 adversarial examples via MUROAN framework and trained the model on these samples combined with the original dataset from scratch. 
We observed that the adversarial trained DMM was still vulnerable to newly crafted decoupled samples, despite the model achieving near 100\% accuracy classifying adversarial examples included in the training set. 
These results demonstrate the difficulty in defending against decoupling attacks using traditional adversarial training. We hope these results inspire further work in increasing the robustness of DMMs. 

\section{Discussion}
\label{sec:discussion}
In this section, we discuss some limitations, potential negative societal impacts, and some future directions of our work. 

In this work, we have focused on DMMs that mainly operate on image and text modalities as inputs. We chose this type of DMMs since it could represent different compositions of inputs (i.e., a continuous input and a discrete input). Our approach however can be generalized to incorporate any other types of DMMs, considering  compositions of other inputs including speech and video modalities. 

Our major research objective is to improve the robustness of DMMs by showing their vulnerabilities to decoupling attacks, so that adversarial attacks on real-world multimodal systems can be mitigated. A potential negative societal impact of our work is that it might be used to craft adversarial attacks against DMMs. 
As future work, we will investigate potential techniques to improve the robustness of DMMs. We hope our findings encourage more researchers to pursue improving the robustness of DMMs.

\section{Conclusion}
In conclusion, we have studied the robustness of DMMs against multimodal decoupling attacks that are aimed at compromising the fusion mechanism of DMMs. 
We have introduced a new framework called MUROAN for studying the robustness of DMMs, which consists of a unified view of the DMMs that exposes the fusion embedding, and an algorithm for decoupling the input modalities. 
Our experiment regarding adversarial training shows that it does not improve the robustness against our decoupling attacks. MUROAN paves the way for studying the robustness of DMMs via decoupling input modalities in the future.

\bibliography{bibtexes}

\newpage


\appendix
\section{Appendix}
\label{appendix}

\subsection{Additional Qualitative Results}
\begin{figure}[h]
\centering
\includegraphics[width=0.8\columnwidth]{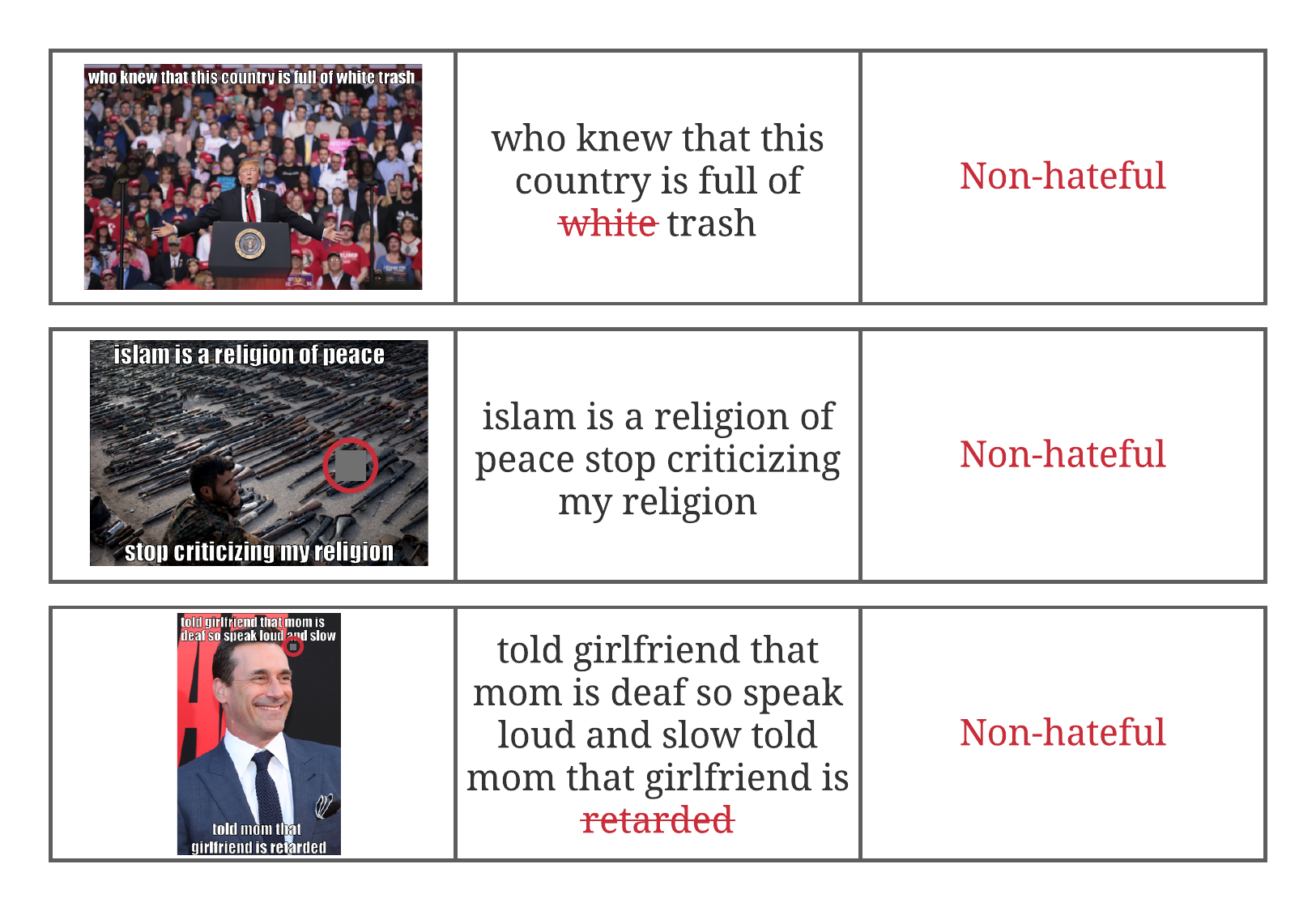} 
\vspace{-0.5cm}
\caption{Additional Samples from the Hateful Memes dataset.}
\vspace{-0.5cm}
\label{fig:qualitative_samples_hm}
\end{figure}

\begin{figure}[h]
\centering
\includegraphics[width=0.8\columnwidth]{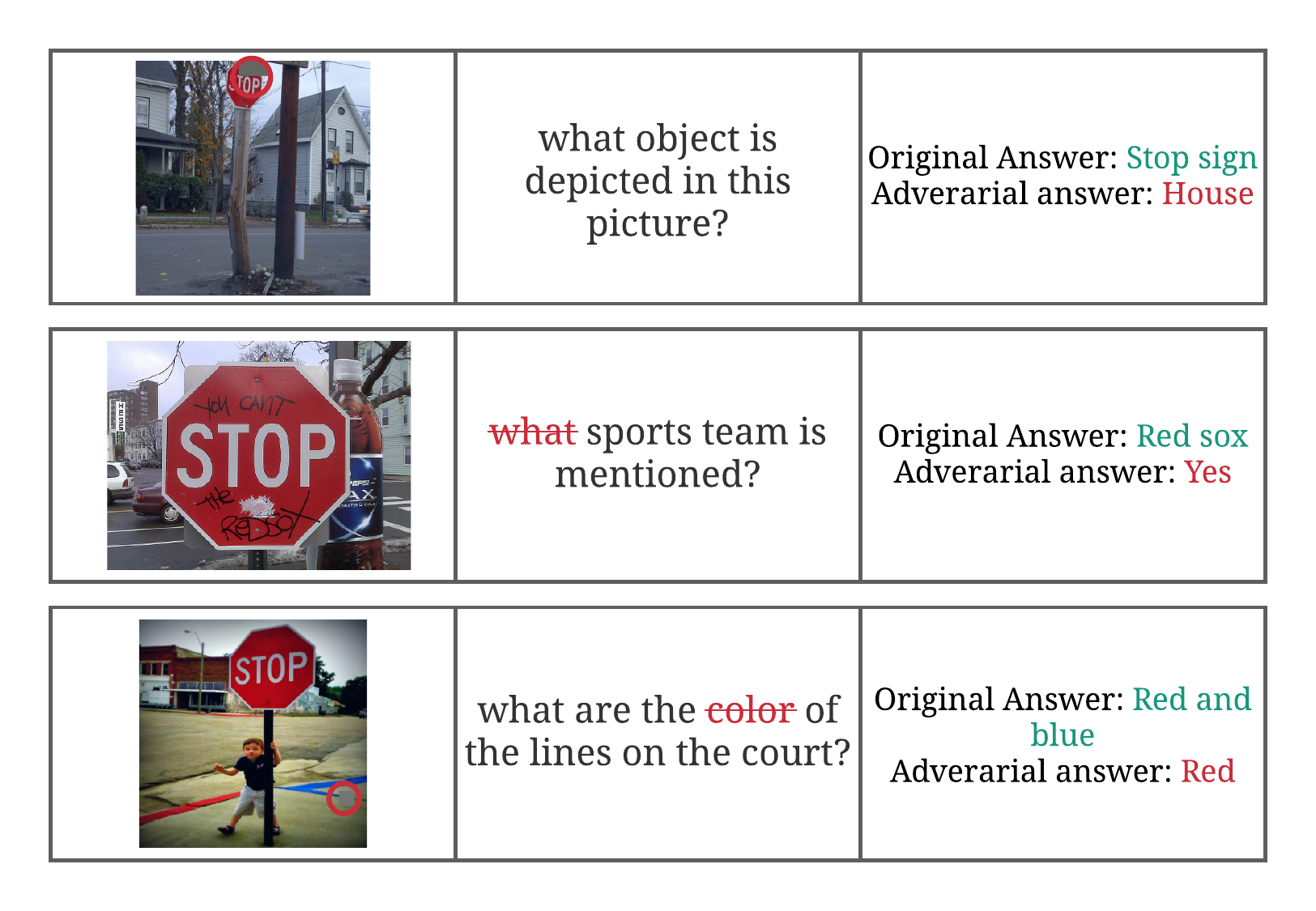} 
\vspace{-0.5cm}
\caption{Additional Samples from the VQA dataset.}
\vspace{-0.5cm}
\label{fig:qualitative_samples_vqa}
\end{figure}

In this section, we provide additional qualitative examples of our attack against the MMBT-Hateful Memes model and the Pythia-VQA model in Figure~\ref{fig:qualitative_samples_hm} and Figure~\ref{fig:qualitative_samples_vqa}, respectively. In Section~\ref{sec:qualitative_analysis_of_mda}, we discussed a few samples from MUROAN from the Hateful Memes dataset. We further discuss more samples from the VQA dataset in addition to some samples from the Hateful Memes dataset in this section.

Figure~\ref{fig:qualitative_samples_hm} depicts three samples from the MMBT-Hateful Memes baseline. The first sample depicts the case where only the text is needed to be manipulated to decouple the input modalities in a sample. The second example depicts the case where only a part of the image needs to be manipulated to decouple the modalities in a sample. The third example depicts the case where both image and the text need to be manipulated to decouple the modalities in a sample.

Figure~\ref{fig:qualitative_samples_vqa} depicts three samples from the Pythia-VQA baseline. In this case, the objective is to fool the DMM so as to output a wrong answer (as opposed to a wrong label in the Hateful Memes case). We observed a similar trend in case of VQA as well, as noted in Section~\ref{sec:qualitative_analysis_of_mda}. In some cases (such as the first sample and the second sample in Figure~\ref{fig:qualitative_samples_vqa}), it was sufficient to only manipulate one of the input modalities to decouple the input modalities in a sample. In some cases though, both modalities had to be manipulated for decoupling them (such as the third sample in Figure~\ref{fig:qualitative_samples_vqa}).

\subsection{Quantitative Robustness Analysis of DMMs}

We have discussed in Section~\ref{sec:Robustness_Analysis} about how our attack can be used to study the robustness of several DMMs. In this section, we use our attack to study and compare the robustness of two baseline DMMs, Late Fusion and MMBT, discussed in our paper. 

\begin{figure}[h]
\vspace{-0.1cm}
    \centering
    \begin{subfloat}[][Late Fusion Model]{
        \includegraphics[width=\linewidth]{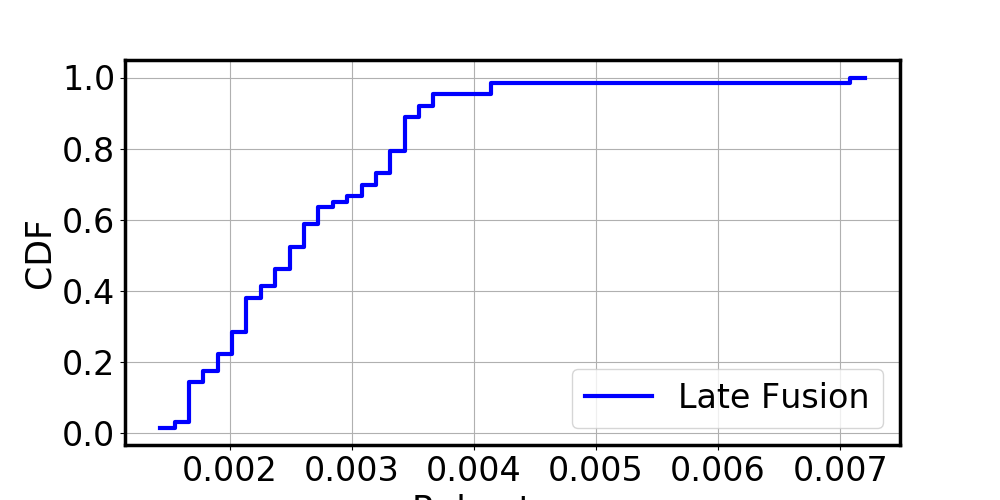}\label{fig:robustness_late_fusion}}
    \end{subfloat}
    ~%
    \begin{subfloat}[][MMBT]{
        \includegraphics[width=\linewidth]{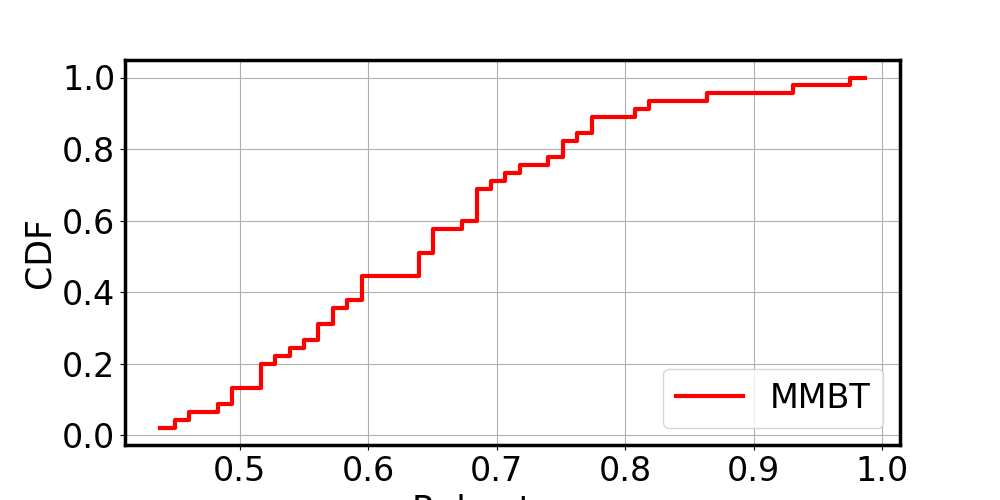}\label{fig:robustness_mmbt}}
    \end{subfloat}
    \caption{CDF of robustness of Late Fusion model and MMBT against MUROAN decoupling attack algorithm.}
    \label{fig:robustness}
    \vspace{-0.2cm}
\end{figure}

 In this experiment,  our objective is to compare the two DMMs that are trained for the same task to determine which DMM is more robust against our attack. In this way, we can use MUROAN to additionally compare DMMs in terms of their robustness. We study and compare the robustness of the two DMMs both trained on the Hateful Memes dataset based on the \textit{robustness metric} $\psi$~\cite{yu2019interpreting}. Model robustness is defined as follows.

\begin{equation}
\label{eq:robustness}
\psi(x) = \frac{1}{\underset{\delta \in set}{\max} D_{KL} (P(x), P(x + \delta))}
\end{equation}

Equation~\ref{eq:robustness} uses the Kullback–Leibler divergence loss ($D_{KL}$)~\cite{kullback1997information} to depict the divergence between the probability distributions of the original samples and the adversarial samples generated by MUROAN decoupling attack algorithm. In other words, the $D_{KL}$ is higher for a model, for which the adversarial samples are further from the original distribution, indicating stronger robustness. In this experiment, we compared the robustness of the MMBT model to the Late Fusion model, where both DMMs were trained on the same Hateful Memes dataset. The distribution of the robustness the two DMMs as calculated by Equation~\ref{eq:robustness} based on our attack is depicted in Figures~\ref{fig:robustness_late_fusion} and~\ref{fig:robustness_mmbt}, respectively. We found that the MMBT model is significantly more robust than the Late Fusion model, as can be observed from the Figure~\ref{fig:robustness}. The mean robustness of the MMBT model was found to be $\psi = 0.65$ and the mean robustness of the Late Fusion model was found to be $\psi = 0.003$. The higher robustness of the MMBT model could be attributed to the way the fusion is achieved in this DMM, using the more sophisticated self-attention mechanism of the  transformer~\cite{vaswani2017attention}, while the Late Fusion model uses the element-wise addition. Thus, the robustness metric in this experiment could also indicate the strength of the fusion mechanism. In this way, the robustness of the  state-of-the-art DMMs can be quantitatively measured using MUROAN.

\end{document}